\documentclass[runningheads]{llncs}
\usepackage{graphicx}
\usepackage{amsmath}
\usepackage[inline]{enumitem}
\usepackage{todonotes}
\usepackage{algorithm}
\usepackage{algorithmicx}
\usepackage{graphicx}
\usepackage{makecell}
\usepackage[colorlinks=true,allcolors=blue]{hyperref}
\usepackage{subfig}
\usepackage{booktabs,multirow}
\usepackage{pgf,tikz,pgfplots}
\usepackage[utf8]{inputenc}
\usepackage{pgfplots}
\usepackage[noend]{algpseudocode}%\DeclareUnicodeCharacter{2212}{−}
\usepgfplotslibrary{groupplots,dateplot}
\usetikzlibrary{patterns,shapes.arrows}
\pgfplotsset{compat=newest}

\newcommand{\flipw}{f}
\newcommand{\dynflipw}{w}

\begin{document}
\title{Towards Tackling MaxSAT by Combining Nested Monte Carlo with Local Search}
%\author{Hui Wang, Abdallah Saffidine, Tristan Cazenave}
\institute{LAMSADE University Paris Dauphine PSL \and The University of New South Wales}
\author{Hui Wang\inst{1} \and
Abdallah Saffidine\inst{2}\and
Tristan Cazenave\inst{1}}
%\institute{Princeton University, Princeton NJ 08544, USA \and
%Springer Heidelberg, Tiergartenstr. 17, 69121 Heidelberg, Germany
\maketitle              % typeset the header of the contribution
\begin{abstract}
Recent work proposed the UCTMAXSAT algorithm to address Maximum Satisfiability Problems (MaxSAT) and shown improved performance over pure Stochastic Local Search algorithms (SLS).
UCTMAXSAT is based on Monte Carlo Tree Search but it uses SLS instead of purely random playouts. In this work, we introduce two algorithmic variations over UCTMAXSAT. We carry an empirical analysis on MaxSAT benchmarks from recent competitions and establish that both ideas lead to performance improvements.
First, a nesting of the tree search inspired by the Nested Monte Carlo Search algorithm is effective on most instance types in the benchmark.
Second, we observe that using a static flip limit in SLS, the ideal budget depends heavily on the instance size and we propose to set it dynamically. We show that it is a robust way to achieve comparable performance on a variety of instances without requiring additional tuning.
\end{abstract}
%\todo{to add dynamic flip limit for nmcts in fig1}
\section{Introduction}
Maximum Satisfiability~(MaxSAT) problem is an extension of Boolean Satisfiability~(SAT) problem. For MaxSAT, the task is to find a truth value assignment for each literal which satisfies the maximum number of clauses~\cite{goffinet2016monte}. Stochastic Local Search~(SLS) algorithms like WalkSat~\cite{kautz2004walksat} and Novelty~\cite{menai2003efficient} are well studied to solve MaxSAT problems. These methods can not find a provable optimal solution but are usually used to search for an approximate optimal solution especially for larger problem instances. However, SLS algorithms are easy to get stuck in a local optimal solution and it's hard for them to escape. Thus, it's important to find an effective way to get rid of the local optimal solution. As a well-known successful method to address this exploration-exploitation dilemma, Monte Carlo Tree Search~(MCTS) with UCT formula~\cite{browne2012survey} is an ideal algorithm to deal with MaxSAT problems.

MCTS has shown impressive performance on game playing~(including perfect information games and imperfect information games)~\cite{gelly2007combining,cowling2012ensemble,wang2018assessing}, probabilistic single-agent planning~\cite{seify2020single}, as well as most of problems which can be formed as a sequential decision making process, also know as Markov Decision Process~(MDP)~\cite{brechtel2011probabilistic}. Based on the UCT formula, MCTS can address the exploration and exploitation dilemma in a theoretically sound way because UCT provides a state-of-the-art way to build the search tree based on the previous search records~(including the node visited count and the node estimate values of the visit). Typically, the estimate 
method of the leaf node in the search tree is a random rollout policy. However, in a lot of applications, many other rollout policies are created to improve the accuracy of the leaf node value estimation. %For examples, in game playing, AlphaGo series programs employ the neural networks to provide the estimate value for the node~(state)~\cite{silver2016mastering,silver2017mastering,silver2018general,wang2020warm,wang2021adaptive}. 
For MaxSAT problem, UCTMAXSAT~(simply written as UCTMAX in the following parts) employs SLS algorithms to estimate the node value~\cite{goffinet2016monte}. %This paper has explored the effect of using random rollout or possible heuristics for MaxSAT, we find that heuristics can improve the performance compared to a random rollout, but SLS based search algorithms are far better than random rollout and the heuristics.

However, UCTMAX only runs MCTS for the root node to build a search tree until the time out, which may not sufficiently use the advantage of UCT reported by the Nested Monte Carlo Tree Search~(NMCTS)~\cite{baier2012nested}. NMCTS runs MCTS from root to the end or the time out. For each step, after performing the MCTS, it chooses the best assignment value for the current step and then enters into the next step and performs the MCTS again. In addition, UCTMAX employs a fixed flip limit for SLS algorithms. But in a UCT-style SLS, the number of the unassigned variables~(literals below the search tree frontier are unassigned) will decreases along with the search tree deepens. Therefore, we design a novel computation called \emph{Dynamic SLS}, see Equation~\ref{eq:dynamicflip}, for Monte Carlo methods used in this paper. %Dynamic SLS uses a weight number $w$ to calculate the flip limit $f$ according to the number of the unassigned variables $v$, see Equation~\ref{eq:dynamicflip}. 
The experimental results show that for most of the MaxSAT instances~\footnote{The instances are from \emph{$ms\_random$} benchmark:\\ \url{http://www.maxsat.udl.cat/15/benchmarks/index.html}}, the Dynamic SLS way is more robust than the fixed way used for UCTMAX  to achieve comparable performance on a variety of instances without extra tuning. Besides, the results show that the NMCTS is better than the UCTMAX on most instances with moderate improvement.

Moreover, Nested Monte Carlo Search~(NMCS) method~\cite{cazenave2009nested} and its variants~\cite{cazenave2020generalized,cazenave2012application,cazenave2020generalized} have been successfully applied to master many NP-hard combinatorial optimization problems, like Morpion Solitaire~\cite{demaine2006morpion}, and achieve impressive performance~\cite{cazenave2009nested,wang2020tackling}. However, NMCS has not been investigated to deal with MaxSAT problems. Therefore, this paper further studies the effectiveness of NMCS~(also using Dynamic SLS as the state estimate) for MaxSAT. %The experimental results show that for most of the MaxSAT instances, WalkSat based NMCS and ZNMCS~(a variant of NMCS, see Sec.~\ref{subsec:nmcs}) outperform NMCTS, and the level 1 is better than level 2 for NMCS, level 2 is better than level 1 for ZNMCS given a fixed time budget. Besides, Novelty based NMCTS outperforms others in larger instances. All Novelty based methods achieve better results than WalkSat based methods.

Overall, the main contribution of this paper can be summarized as follows:

\begin{enumerate}

    \item We examine various Monte Carlo Search techniques for the domain of MaxSAT, especially rollout policies and high-level searches.
    Through an extensive empirical analysis, we establish that
    \begin{enumerate*}
    \item Purely random or heuristic-based rollouts are weaker than a Stochastic Local Search policy.
    \item An MCTS-based search is weaker than Nested MCTS, especially in larger instances. NMCTS with WalkSat is weaker than NMCS, but is stronger with Novelty.
    \end{enumerate*}
    \item We introduce Dynamic SLS, a new rollout policy that dynamically computes the flip budget available for a stochastic local search. We demonstrate that Monte Carlo algorithms building on Dynamic SLS achieve comparable performance on standard MaxSAT benchmarks with previously existing Monte Carlo approaches without extra tuning.
\end{enumerate}

%to merge section 6 to section 4. 
%to say in text that global 

The rest of the paper is structured as follows. 
Before introducing preliminaries of this work in Sec.~\ref{sec:preliminaries}, we present an overview of the most relevant literature in Sect.\,\ref{sec:relatedwork}. Then we present Dynamic SLS based Monte Carlo methods in Sect.\,\ref{sec:slsmctsnmc}. Thereafter, we illustrate the orientation experiments on a group of MaxSAT instances to finalize the structure of our proposed methods in Sect.\,\ref{sec:orientation}. Then the full length experiments are presented in Sect.\,\ref{sec:full-exp}. %In addition, in Sect.\,\ref{trialplayout}, we test different playout policy for MCTS and NMCS and present the light experimental results. 
Finally, we conclude our paper and discuss future work.

\section{Related Work}\label{sec:relatedwork}
%MaxSat problem is a generation of SAT problem and typically a class of NP-hard problem. The goal of MaxSat is to assign literals with different true or false values so that the assignment values satisfy the maximun number of clauses~\cite{goffinet2016monte}. 
There are a lot of solvers created to master MaxSAT problems~\cite{heras2008minimaxsat,martins2014open,ansotegui2017wpm3,ignatiev2019rc2}. Generally, these solvers can be categorized into two different types, i.e. $complete$ solvers and $incomplete$ solvers. Complete solvers can provide provable the best solution for the problem. Incomplete solvers start from a random assignment and continue to search for a better solution according to some strategies. Typically, Stochastic Local Search algorithms like WalkSat~\cite{kautz2004walksat} and Novelty~\cite{menai2003efficient} are well studied on MaxSAT~\cite{pelikan2003hierarchical,kroc2009integrating}. These $incomplete$ solvers suffer from an exploration-exploitation dilemma. And MCTS has shown successful performance of dealing with this dilemma~\cite{browne2012survey}. Therefore, Tompkins et al. implemented an experimentation environment for mastering SAT and MaxSAT, called UCBMAX~\cite{tompkins2004ubcsat}. Furthermore, Goffinet et al proposed UCTMAX algorithm to enhance the performance of SLS~\cite{goffinet2016monte}. However, UCTMAX only performs UCT search once from the root, which 
may not sufficiently use the power of MCTS comparing to run UCT search for each step until to the terminal node or time out, which is known as Nested Monte Carlo Tree Search~\cite{baier2012nested}. In addition to MCTS and NMCTS, NMCS~\cite{cazenave2009nested} and its variations~\cite{cazenave2012application,rosin2011nested,cazenave2020generalized} also perform well especially for single agent NP-hard combinatorial problems, like Morpion Solitaire~\cite{demaine2006morpion}, where they achieve the best record which has not yet been improved even employing deep learning techniques~\cite{wang2020tackling,douxdeep}. Therefore, in this paper, we firstly employ NMCTS and NMCS to master MaxSAT problems with SLS methods. %Meanwhile, we also test NMCS, NMCTS with random playout instead of any specific state estimation heuristics.

\section{Preliminaries}\label{sec:preliminaries}
\subsection{MaxSAT}
 In MaxSAT, like SAT, the problem is specified by a propositional formula described in conjunctive normal form (CNF)~\cite{morgado2013iterative}. But unlike SAT which the aim is to find a truth assignment to satisfy all clauses, MaxSAT is just to find a truth assignment to satisfy the maximum number of clauses. For a set of Boolean variables $V=\{v_1, v_2, v_3, ..., v_i\}$, a literal $l_j$
is either a variable $v_j$ or its negation $\neg v_j$, $1\leq j\leq i$. A clause is a disjunction of literals (i.e., $c_i = l_1 \vee l_2 \vee, ...,  \vee l_j$ ). A CNF formula $F$ is a set of clauses as conjunctive normal form (i.e., $F = c_1\wedge c_2 \wedge, ..., \wedge c_i$). MaxSAT instances written as CNF can be easily found in our tested benchmark. %We demonstrate an example of CNF file, which indicates this MaxSAT instance has 3 literals and 4 clauses, for each clause, a 0 is used as separation. 

%c comments Max-SAT

%p cnf 3 4

%1 -2 0

%-1 2 -3 0

%-3 2 0

%1 3 0

\subsection{Heuristics}
In order to test the different rollout policies for Monte Carlo Methods, here we present 3 simple heuristics that commonly used for MaxSAT.
\begin{enumerate}
    \item H1 is the heuristic which assigns the value \emph{from the first variable to the last variable} and H1 sets 0 for a variable that its positive value occurs more times than its negative value in all clauses.
    \item H2 is the heuristic which, for each step, assigns the \emph{variable} first which occurs the most times and H2 sets 0 for a variable that its positive value occurs more times than its negative value in all clauses.
    \item H3 is the heuristic which, for each step, assigns the \emph{literal} first which occurs the most times and H3 sets 0 for a variable that its positive value occurs more times than its negative value in all clauses.
\end{enumerate}

\subsection{Stochastic Local Search}
Based on~\cite{goffinet2016monte}, in this paper, we also investigate two well-studied Stochastic Local Search~(SLS) algorithms to deal with MaxSAT problem, namely WalkSat and Novelty.
\subsubsection{WalkSat}\label{subsec:walksat}
\begin{algorithm}[bth!]
\caption{Walksat}
\label{alg:walksat}
\begin{algorithmic}[1]
%\footnotesize
%\Function{Walksat}{s}
%\State InitAssignment()
%\While {fliptimes $< f$}
%\State get a random or a best position $pos$ in an unsatisfied clause based on an $\epsilon$ probability
%\State flip(pos)
%\EndWhile
%\EndFunction
%\Function{InitAssignment}{}
%\State Init each unAssigned literal $lit$ according to the best solution found so far with another $\epsilon$ probability
%\EndFunction
%\Function {getBestPositiontoFlip}{}
%\For{each literal $lit$ in the selected False clause}
%\State Compute the bonus of flipping $lit$
%\State pos=$lit$ with the highest bonus
%\EndFor
%\EndFunction
\Function {WalkSat}{$s$}
%\State declare $v_f$
\State assignment$\leftarrow$\textsc{InitAssignment}()
\While {fliptimes $< f$}
\If{\textsc{random}() $< \epsilon_1$}
\State $v \leftarrow$ random variable
\Else
\State $v \leftarrow$ best unassigned variable\label{bestvariblefound}
%\If{$v = v_f 9$ and \textsc{random}() $< 1-\epsilon_2$}
%\State $v \leftarrow$ second best unassigned variable
%\EndIf
\EndIf
\State assignment$\leftarrow$flip($v$)
%\State $v_f \leftarrow v$
\EndWhile
\Return assignment
\EndFunction
\end{algorithmic}
\end{algorithm}
As it can be seen in Algorithm~\ref{alg:walksat}, the idea of WalkSat is to initialize a random assignment~(basic version) or according to the current found best solution~(enhanced version) for each variable. Then an unsatisfied clause is selected. Further step is to select a variable to flip which has the highest bonus after flipping in the selected unsatisfied clause. The bonus is the change of the number of satisfied clauses after flipping the variable.

\subsubsection{Novelty}\label{subsec:novelty}
%\begin{algorithm}[bth!]
%\caption{Novelty}
%\label{alg:novelty}
%\begin{algorithmic}[1]
%\footnotesize
%\Function {Novelty}{$s$}
%\State most recent flipped variable $v_f$,
%\State \textsc{InitAssignment}()
%\While {fliptimes $< f$}
%\State with $\epsilon_1$ probability $v \leftarrow$\textsc{RandomVariable}() else $v \leftarrow$ \textsc{getBestVariabletoFlip}()
%\State flip($v$)
%\State $v_f \leftarrow v$
%\EndWhile
%\EndFunction
%\Function{getBestVariabletoFlip}{}
%\State $v, v_2 \leftarrow$ best and second best variables among all unassigned variables
%\If{$v= v_f$}
%\State with probability $1-\epsilon_2$ set $v \leftarrow v_2$
%\EndIf
%\Return $v$
%\EndFunction

%\Function {Novelty}{$s$}
%\State declare $v_f$
%\State \textsc{InitAssignment}()
%\While {fliptimes $< f$}
%\If{\textsc{random}() $< \epsilon_1$}
%\State $v \leftarrow$ random variable
%\Else
%\State $v \leftarrow$ best unassigned variable
%\If{$v = v_f$ and \textsc{random}() $< 1-\epsilon_2$}
%\State $v \leftarrow$ second best unassigned variable
%\EndIf
%\EndIf
%\State flip($v$)
%\State $v_f \leftarrow v$
%\EndWhile
%\EndFunction
%\end{algorithmic}
%\end{algorithm}

Novelty is similar to WalkSat. The first step is also to initialize a random assignment~(basic version) or according to the current found best solution~(enhanced version). But differently, for each variable in all unsatisfied clauses, its bonus is computed. Then in order to avoid flipping in a dead loop, a variable which has the highest bonus but not selected in the most recent flipping is selected to flip. Simply, after line~\ref{bestvariblefound} in Algorithm~\ref{alg:walksat}, we add \textbf{If $v = v_f$ and \textsc{random}() $< 1-\epsilon_2$ then $v \leftarrow v_s$}. $v_f$ is the most recent flipped variable and $v_s$ is the second best unassigned variable.

\subsection{Monte Carlo Tree Search}
\begin{algorithm}[bth!]
\caption{Monte Carlo Tree Search}
\label{alg:mcts}
\begin{algorithmic}[1]
\footnotesize
\Function{MCTS}{$s$}
\State Search($s$)
\State $\pi_s\leftarrow$normalize($Q(s,\cdot)$)\label{line:getpolicy}
\State \Return $\pi_s$ 
\EndFunction
\Function{Search}{$s$}
\If{$s$ is a terminal state}
\State $v\leftarrow v_{end}$\label{mctsvalueend}
\Return $v$ 
\EndIf
\If{$s$ is not in the Tree}
\State Add $s$ to the Tree, initialize $Q(s, \cdot)$ and $N(s, \cdot)$ to 0
\State Run rollout policy and get the solution score $v_{rollout}$\label{mctsrollout}
\State $v\leftarrow v_{rollout}$\label{mctsvaluerollout}
\Return $v$
\Else
\State Select an action $a$  with highest UCT value\label{line:uct}
\State $s^\prime\leftarrow$getNextState($s$, $a$)
\State $v\leftarrow$Search($s\prime$)
\State $Q(s,a)\leftarrow\frac{N(s,a)*Q(s,a)+v}{N(s,a)+1}$
\State $N(s,a)\leftarrow N(s,a)+1$ \label{line:update}
\EndIf
\State \Return $v$
\EndFunction
\end{algorithmic}
\end{algorithm}
According to~\cite{wang2020analysis,wang2020warm,wang2021adaptive}, a recursive MCTS pseudo code is given in Algorithm~\ref{alg:mcts}. 
For each search, the rollout value is returned (or the game termination score). For each visit of a non-leaf node, the action with the highest UCT value is selected to investigate next~\cite{browne2012survey}. 
After each search, the average win rate value $Q(s,a)$ and visit count $N(s,a)$ for each node in the visited trajectory is updated correspondingly. The UCT formula is as follows: 
\begin{equation}
   U(s,a) = Q(s,a) +c \sqrt{\frac{ln(N(s,\cdot))}{N(s,a)+1}}
\end{equation}

The Nested Monte Carlo Tree Search~(Due to the high computation, we only investigate level 1 for NMCTS in this paper) calls MCTS for each step of the assignment process.

\subsection{Nested Monte Carlo Search}\label{subsec:nmcs}
\begin{algorithm}[bth!]
\caption{Nested Monte Carlo Search}
\label{alg:nmc}
\begin{algorithmic}[1]
\footnotesize
\Function{NMC}{$s$, level}
\State chosenSeq$\leftarrow$[], bestScore$\leftarrow -\infty$, bestSeq$\leftarrow$[]
\While{$s$ is not terminal} 
\For {each $m$ in legalMoves($s$)}\label{alg:nmc:linefor}
\State $s^\prime\leftarrow$ PerformMove($s$, $m$)\label{alg:nmc:lineperform}
\If{level = 1}
\State (score, seq) $\leftarrow$ run rollout policy\label{nmcrollout}
\Else
\State (score, seq) $\leftarrow$  NMC($s^\prime$, level-1)\label{changetozmcs}
\EndIf
\EndFor
\State highScore $\leftarrow$  highest score of the moves from $s$
\If {highScore $>$ bestScore}
\State bestScore $\leftarrow$  highScore
\State chosenMove $\leftarrow$ $m$ associated with highScore
\State bestSeq $\leftarrow$  seq associated with highScore
\Else
\State chosenMove $\leftarrow$  first move in bestSeq
\State bestSeq $\leftarrow$  remove first move from bestSeq
\EndIf
\State $s \leftarrow$  perform chosenMove to $s$
\State chosenSeq $\leftarrow$  append chosenMove to chosenSeq
\EndWhile
\State \Return (bestScore, chosenSeq);
\EndFunction
\end{algorithmic}
\end{algorithm}

According to~\cite{cazenave2009nested}, the Nested Monte Carlo Search algorithm employs nested calls with rollouts and the record of the best sequence of moves with different levels. The basic
level only performs random moves. Since a nested search may obtain worse results than a previous lower level
search, recording the currently found best sequence and following it when the searches result in
worse results than the best sequence is important. Therefore, we present the pseudo code for the basic Monte Carlo Search algorithm as~\ref{alg:nmc}. In order to estimate the leaf nodes from themselves instead of their children, we further test a variant of NMCS, named ZNMCS~(Zero Nested Monte Carlo Search), where in Algorithm~\ref{alg:nmc}, line~\ref{alg:nmc:linefor} is changed to \textbf{for $i=0, i<t, i++$ do}, in our experiments, $t=10$. In addition, line~\ref{alg:nmc:lineperform} has been removed. And line~\ref{changetozmcs} is changed to (score, seq)$\leftarrow$ ZNMCS($s$, level-1).

\section{Dynamic SLS Based Monte Carlo Methods}\label{sec:slsmctsnmc}
%\subsection{Dynamic SLS}\label{dynsls}
This section proposes the Dynamic SLS method with MCTS and NMCS.
Since the number of the unassigned variables decreases as the search tree deepens, we propose a Dynamic SLS to avoid redundant flips and enlarge search tree to improve the performance within a fixed time budget.
%The dynamic decreasing flip limit~(written as $\flipw$) is simply computed according to the following Equation: 
The flip limit~(written as $\flipw$) is simply computed according to the following Equation: 
\begin{equation}\label{eq:dynamicflip}
    \flipw=\dynflipw \times u
\end{equation}    
$w$ is a weight number, $u$ is the number of the unassigned variables which can be flipped. Considering MCTS, in the search tree, the variables, upon the leaf nodes, have already been assigned to a value, so they can not be flipped anymore. We also tested several exponent values powered by $u$ and finally found exponent equals 1 is the best.

In this work, we insert Dynamic SLS to replace rollout policy for MCTS~(line~\ref{mctsrollout} in Algorithm~\ref{alg:mcts}) and NMCS~(line~\ref{nmcrollout} in Algorithm~\ref{alg:nmc}, same to ZNMCS). In addition, according to~\cite{goffinet2016monte}, it is reported that using square number of the score is the best for UCTMAX, so in this work, for MCTS, we also replace the value calculation in line~\ref{mctsvalueend} and line~\ref{mctsvaluerollout} in Algorithm~\ref{alg:mcts} to $v=pow(v_{end}, 2)$ and $v=pow(v_{dsls}, 2)$.

%\subsection{Dynamic SLS Based Monte Carlo Methods}\label{subsec:dmcm}

\section{Orientation Experiments}\label{sec:orientation}

\subsection{Trial with Different Rollout}\label{subsect:trialplayout}
There are several ways to estimate the state value for Monte Carlo methods. One typical way is to simply run random simulations to get approximate values. In addition, for MaxSAT, there are many well designed heuristics to assign the truth values, based on the assignment, a proper value can be obtained. Besides, there are also several well studied SLS algorithms which can be applied to estimate the state value. Therefore, in order to determine which way is the best for the state estimate function, we use different ways to work together with NMCTS and NMCS to process our test setting~(50 different instances, 70 variables each). The NMCTS simulation is set as 100. Time cost for each run is 50 seconds. each setting runs 10 repetitions. The results are shown in Table~\ref{tab:rolloutpolicies}. We see that the heuristics all outperform random rollout, H3 is better than H2 and H2 is better than H1. Importantly, SLS methods perform significantly the best. So we adopt WalkSat and Novelty as the rollout policies for the further experiments. In addition, WalkSat for NMCS is better than NMCTS, but NMCTS with Novelty is the best.

\begin{table}[bht!]
\caption{Results for Max3Sat Instances~(70 variables) Using Different Rollout Policies for MCTS, NMCS.
Results are average number of unsatisfied clauses on tested group instances, same to the following results.}
\centering
\begin{tabular}{l*1r*3r}
\toprule
Method & NMCTS &  \multicolumn{3}{c}{NMCS}\\
\cmidrule(lr){3-5}
Level & - & playout & level 1 & level 2 \\
\midrule
Random&81.4 &125.2 &80.8 &80.5\\
H1&56.1 &70.0& 54.4& 53.7\\
H2&55.1 & 69.5 &54.6 &53.8\\
H3&53.2& 64.4 &52.2 &52.2 \\
WalkSat&47.9&52.0&\textbf{47.4}&\textbf{47.7}\\
Novelty&\textbf{47.7}&\textbf{51.9}&48.8&49.0\\
\bottomrule
\end{tabular}\label{tab:rolloutpolicies}
\end{table}

\subsection{UCTMAX vs NMCTS}\label{subsect:SingleUCTvsMultiUCT}
Since \cite{goffinet2016monte}
only investigated the UCTMAX with one time MCTS from the root until the time out. However, it does not perform an action to enter next state and run UCT again like game playing. To this end, the NMCTS~\cite{baier2012nested} method should be further investigated. We let the MCTS simulation as a fixed value~(set as 100) so that each step will stop and get a search tree. Based on this search tree, a best action can be found and performed to enter to next state. Then it runs another UCT process until the time out or the termination. The results show that the NMCTS performs clearly better than the UCTMAX way. In order to enlarge the result difference for different settings, we use larger instances~(50 instances, each has 140 variables. \cite{goffinet2016monte} also used 140 as the test instance size, but they only tested on one instance, we test on 50 different instances with this size to reduce the noisy.) for this experiment and the following orientation experiments.

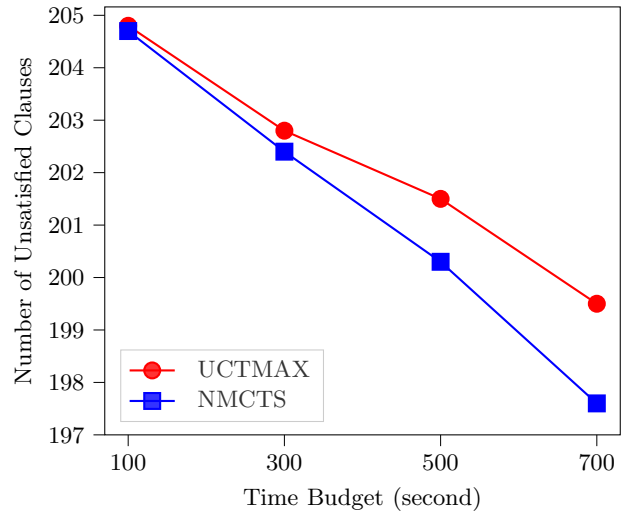
\begin{figure}[bth!]
\centering
\hspace*{-1.5em}
% This file was created with tikzplotlib v0.10.1.
\begin{tikzpicture}

\definecolor{darkgray176}{RGB}{176,176,176}
\definecolor{lightgray204}{RGB}{204,204,204}

\begin{axis}[
legend cell align={left},
legend style={
  fill opacity=0.8,
  draw opacity=1,
  text opacity=1,
  at={(0.03,0.03)},
  anchor=south west,
  draw=lightgray204
},
minor xtick={},
minor ytick={},
tick align=outside,
tick pos=left,
x grid style={darkgray176},
xlabel={Time Budget (second)},
xmin=70, xmax=730,
xtick style={color=black},
xtick={100,300,500,700},
y grid style={darkgray176},
ylabel={Number of Unsatisfied Clauses},
ymin=197, ymax=205.16,
ytick style={color=black},
ytick={197,198,199,200,201,202,203,204,205}
]
\addplot [thick, red, mark=*, mark size=3, mark options={solid}]
table {%
100 204.8
300 202.8
500 201.5
700 199.5
};
\addlegendentry{UCTMAX}
\addplot [thick, blue, mark=square*, mark size=3, mark options={solid}]
table {%
100 204.7
300 202.4
500 200.3
700 197.6
};
\addlegendentry{NMCTS}
\end{axis}

\end{tikzpicture}
\caption{Comparison of UCTMAX with NMCTS. NMCTS outperforms UCTMAX on 50 instances which has 140 variables each. For both UCTMAX and NMCTS, the $\flipw$ is set as 2000 which is reported as the best.}
\label{fig:orientation_singlemulti_walksat} 
\end{figure} 

\subsection{Current Global Best Solution}\label{subsect:globalbest}
%to mark the xticks
Based on~\cite{goffinet2016monte} and \cite{cazenave2009nested}, we know that it is the key to keep the global best solution~(the best of the local solutions from all steps) found so far and initialize the SLS algorithms with this global best solution. We still do not know whether it is also important in our Nested Monte Carlo Methods with SLS. Therefore, we design different combinations to show the importance. 

\begin{table}[bth!]
\caption{Impact of Random variable initialization and of keeping the global best solution on the performance of NMCTS and NMCS.
Fixed number of flips (2000), 50 instances, 140 variables each.}\label{tab:currentbest}
\centering
\begin{tabular}{l*5r}
\toprule

Keep Global & \multicolumn{2}{c}{No} & \multicolumn{3}{c}{Yes} \\
\cmidrule(lr){2-3}\cmidrule(lr){4-6}
Initialization & Rand & Best & Rand &Local& Best \\
\midrule
Time Budget & \multicolumn{5}{c}{100s}\\
\cmidrule(lr){2-6}
NMCTS  & 221.2 & 220.8 & 204.8 &205.1& \textbf{204.6} \\
NMCS  & 198.8 & 199.1 & 199.3 &198.8& \textbf{198.7} \\
\toprule
Time Budget & \multicolumn{5}{c}{300s}\\
\cmidrule(lr){2-6}
NMCTS  & 219.9 & 219.8 & 202.9 &202.9 & \textbf{202.6}\\
NMCS  & 195.3 & 195.6 & 195.3 &195.6& \textbf{193.1}\\
\bottomrule
\end{tabular}
\end{table}

The results are shown in Table~\ref{tab:currentbest}, we see that with a small time budget~(100 seconds), for NMCTS, keeping the global best records has shown the advantage, and initializing based on the global best records is also better than not but with small improvements. For NMCS, with 100 seconds, although we still find that keeping the global best records and initializing with them is the best, but it's not very significant. However, we see a clear improvement with larger time budget~(300 seconds). The reason that different initialization does not differ too much might be that the flip limit is set too big so even if it is initialized from random, it can also reach a global record level after flipping. From this experiment, we can conclude that keeping the global best records and initializing based on them for SLS~(in this case, it is WalkSat) are both important to the nested search. NMCS works better than NMCTS with WalkSat on 140 variables instances.

\subsection{Probabilistic SLS Initialization}\label{subsect:slsinitial} 
%\todo[]{more instances for $\epsilon$ idea， 140, 50 instances for all orientation experiments. lower is better as website, table, graph should use the same measurement}

In order to further investigate the contribution of initializing WalkSat based on the global best solution found so far, we adopt the simplest but commonly used way to balance the exploration and exploitation, $\epsilon$-greedy, to initialize the assignment. 

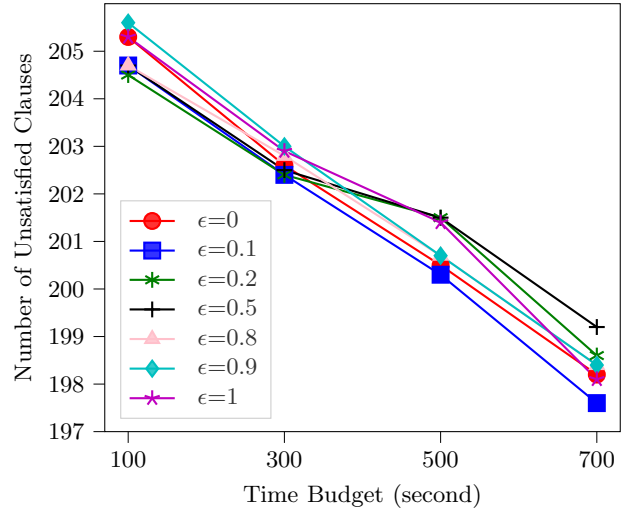
\begin{figure}[bth!]
\centering
\hspace*{-1.5em}
% This file was created with tikzplotlib v0.10.1.
\begin{tikzpicture}

\definecolor{darkgray176}{RGB}{176,176,176}
\definecolor{darkturquoise0191191}{RGB}{0,191,191}
\definecolor{darkviolet1910191}{RGB}{191,0,191}
\definecolor{green01270}{RGB}{0,127,0}
\definecolor{lightgray204}{RGB}{204,204,204}
\definecolor{pink}{RGB}{255,192,203}

\begin{axis}[
legend cell align={left},
legend style={
  fill opacity=0.8,
  draw opacity=1,
  text opacity=1,
  at={(0.03,0.03)},
  anchor=south west,
  draw=lightgray204
},
minor xtick={},
minor ytick={},
tick align=outside,
tick pos=left,
x grid style={darkgray176},
xlabel={Time Budget (second)},
xmin=70, xmax=730,
xtick style={color=black},
xtick={100,300,500,700},
y grid style={darkgray176},
ylabel={Number of Unsatisfied Clauses},
ymin=197, ymax=206,
ytick style={color=black},
ytick={197,198,199,200,201,202,203,204,205}
]
\addplot [thick, red, mark=*, mark size=3, mark options={solid}]
table {%
100 205.3
300 202.6
500 200.5
700 198.2
};
\addlegendentry{ $\epsilon$=0}
\addplot [thick, blue, mark=square*, mark size=3, mark options={solid}]
table {%
100 204.7
300 202.4
500 200.3
700 197.6
};
\addlegendentry{ $\epsilon$=0.1}
\addplot [thick, green01270, mark=asterisk, mark size=3, mark options={solid}]
table {%
100 204.5
300 202.4
500 201.5
700 198.6
};
\addlegendentry{ $\epsilon$=0.2}
\addplot [thick, black, mark=+, mark size=3, mark options={solid}]
table {%
100 204.7
300 202.5
500 201.5
700 199.2
};
\addlegendentry{ $\epsilon$=0.5}
\addplot [thick, pink, mark=triangle*, mark size=3, mark options={solid}]
table {%
100 204.7
300 202.8
500 200.7
700 198.4
};
\addlegendentry{ $\epsilon$=0.8}
\addplot [thick, darkturquoise0191191, mark=diamond*, mark size=3, mark options={solid}]
table {%
100 205.6
300 203
500 200.7
700 198.4
};
\addlegendentry{ $\epsilon$=0.9}
\addplot [thick, darkviolet1910191, mark=star, mark size=3, mark options={solid}]
table {%
100 205.3
300 202.9
500 201.4
700 198.1
};
\addlegendentry{ $\epsilon$=1}
\end{axis}

\end{tikzpicture}
\caption{Initializing Walksat Based on $\epsilon$-greedy for NMCTS on 50 instances with 140 variables each, $\epsilon$=0 means initializing WalkSat totally based on the global best solution. $\epsilon$=0.1 means there is 10\% probability to take a random initialization for the literal, and so on. The $\epsilon$ equals 0.1 is the best.}
\label{fig:orientation_nest_walksat_epsilon}
\end{figure}

From Fig~\ref{fig:orientation_nest_walksat_epsilon}, we see that $\epsilon$=0.1 performs best, which further shows the best initialization way is to set literal assignment based on the best solution found so far but with a small randomness to initialize randomly. Thus, our following experiments are done with the $\epsilon$ as 0.1.

\subsection{Fixed Flip Limits vs Dynamic Flip Limits}
Goffinet et al.~\cite{goffinet2016monte} used the fixed flip limits, which we found can be implemented in a dynamical way. Therefore, in this section, we test different $w$ values~(from 0.5 to 25, but finally we only present results of $w\in \{$1, 2, 4$\}$ as they are better) for dynamic flip limits calculation equation~(see Equation~\ref{eq:dynamicflip}). And we found generally for both NMCTS and NMCS with different budget, $w$=2 is the best~(only the result of 300 second is weaker for NMCTS). In addition, we test fixed flip limit with 2000~(which is reported the best for UCTMAX tuned on a single instance) and 140~(same as the average flip limit for each step with $w$=2). We found that with a fixed flip limit as 2000 is the worst and smaller limits increase the performance which shows that for Nested Monte Carlo methods, allocating time cost for relatively more steps contributes more.

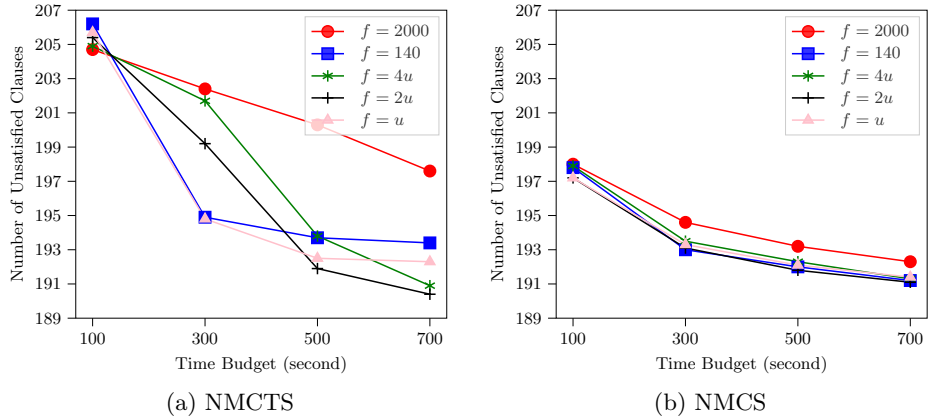
\begin{figure}[!tbh]
\centering
\hspace{-1.5em}
\subfloat[NMCTS]{\label{subfig:orientation_mcts_fixedvsdyn_walksat}
\resizebox{0.49\columnwidth}{!}{% This file was created with tikzplotlib v0.10.1.
\begin{tikzpicture}

\definecolor{darkgray176}{RGB}{176,176,176}
\definecolor{green01270}{RGB}{0,127,0}
\definecolor{lightgray204}{RGB}{204,204,204}
\definecolor{pink}{RGB}{255,192,203}

\begin{axis}[
legend cell align={left},
legend style={fill opacity=0.8, draw opacity=1, text opacity=1, draw=lightgray204},
minor xtick={},
minor ytick={},
tick align=outside,
tick pos=left,
x grid style={darkgray176},
xmin=70, xmax=730,
xtick style={color=black},
xtick={100,300,500,700},
y grid style={darkgray176},
ylabel={Number of Unsatisfied Clauses},
xlabel={Time Budget (second)},
ymin=189, ymax=207,
ytick style={color=black},
ytick={189,191,193,195,197,199,201,203,205,207}
]
\addplot [thick, red, mark=*, mark size=3, mark options={solid}]
table {%
100 204.7
300 202.4
500 200.3
700 197.6
};
\addlegendentry{$\flipw=2000$}
\addplot [thick, blue, mark=square*, mark size=3, mark options={solid}]
table {%
100 206.2
300 194.9
500 193.7
700 193.4
};
\addlegendentry{$\flipw=140$}
\addplot [thick, green01270, mark=asterisk, mark size=3, mark options={solid}]
table {%
100 204.9
300 201.7
500 193.8
700 190.9
};
\addlegendentry{$\flipw=4u$}
\addplot [thick, black, mark=+, mark size=3, mark options={solid}]
table {%
100 205.4
300 199.2
500 191.9
700 190.4
};
\addlegendentry{$\flipw=2u$}
\addplot [thick, pink, mark=triangle*, mark size=3, mark options={solid}]
table {%
100 205.7
300 194.8
500 192.5
700 192.3
};
\addlegendentry{$\flipw=u$}
\end{axis}

\end{tikzpicture}}
}
\hfill
%\hspace{-1.5em}
\subfloat[NMCS]{\label{subfig:orientation_nmcs_fixedvsdyn_walksat}
\resizebox{0.49\columnwidth}{!}{% This file was created with tikzplotlib v0.10.1.
\begin{tikzpicture}

\definecolor{darkgray176}{RGB}{176,176,176}
\definecolor{green01270}{RGB}{0,127,0}
\definecolor{lightgray204}{RGB}{204,204,204}
\definecolor{pink}{RGB}{255,192,203}

\begin{axis}[
legend cell align={left},
legend style={fill opacity=0.8, draw opacity=1, text opacity=1, draw=lightgray204},
minor xtick={},
minor ytick={},
tick align=outside,
tick pos=left,
x grid style={darkgray176},
xlabel={Time Budget (second)},
xmin=70, xmax=730,
xtick style={color=black},
xtick={100,300,500,700},
y grid style={darkgray176},
ymin=189, ymax=207,
ylabel={Number of Unsatisfied Clauses},
ytick style={color=black},
ytick={189,191,193,195,197,199,201,203,205,207}
]
\addplot [thick, red, mark=*, mark size=3, mark options={solid}]
table {%
100 198
300 194.6
500 193.2
700 192.3
};
\addlegendentry{$\flipw=2000$}
\addplot [thick, blue, mark=square*, mark size=3, mark options={solid}]
table {%
100 197.8
300 193
500 192
700 191.2
};
\addlegendentry{$\flipw=140$}
\addplot [thick, green01270, mark=asterisk, mark size=3, mark options={solid}]
table {%
100 197.9
300 193.5
500 192.3
700 191.3
};
\addlegendentry{$\flipw=4u$}
\addplot [thick, black, mark=+, mark size=3, mark options={solid}]
table {%
100 197.2
300 193.1
500 191.8
700 191.1
};
\addlegendentry{$\flipw=2u$}
\addplot [thick, pink, mark=triangle*, mark size=3, mark options={solid}]
table {%
100 197.2
300 193.3
500 192.1
700 191.4
};
\addlegendentry{$\flipw=u$}
\end{axis}

\end{tikzpicture}}
}
\caption{Comparison of Fixed SLS with Dynamic SLS for NMCTS and NMCS. In order to keep the $\dynflipw$ consistent for all runs, considering the overall results, we decide setting the weight $\dynflipw$ for Dynamic SLS flip limits as 2 is the best.
}
\label{fig:orientation_fixedvsdyn_walksat} 
\end{figure} 

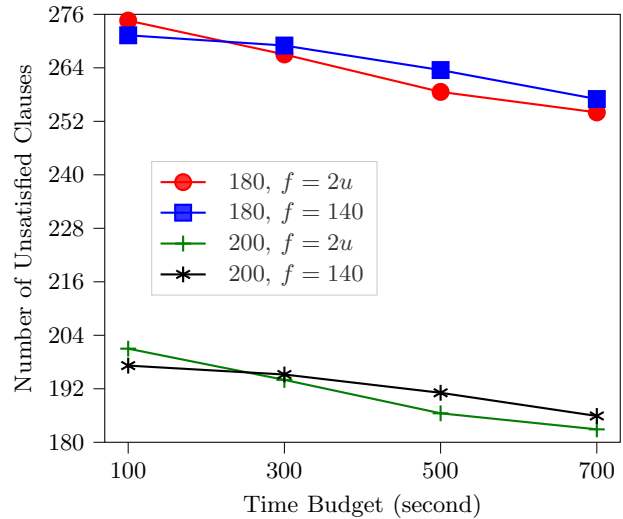
\begin{figure}[bth!]
\centering
\hspace*{-1.5em}
% This file was created with tikzplotlib v0.10.1.
\begin{tikzpicture}

\definecolor{darkgray176}{RGB}{176,176,176}
\definecolor{green01270}{RGB}{0,127,0}
\definecolor{lightgray204}{RGB}{204,204,204}

\begin{axis}[
legend cell align={left},
legend style={
  fill opacity=0.8,
  draw opacity=1,
  text opacity=1,
  at={(0.09,0.5)},
  anchor=west,
  draw=lightgray204
},
minor xtick={},
minor ytick={},
tick align=outside,
tick pos=left,
x grid style={darkgray176},
xlabel={Time Budget (second)},
xmin=70, xmax=730,
xtick style={color=black},
xtick={100,300,500,700},
y grid style={darkgray176},
ylabel={Number of Unsatisfied Clauses},
ymin=180, ymax=276,
ytick style={color=black},
ytick={180,192,204,216,228,240,252,264, 276}
]
\addplot [thick, red, mark=*, mark size=3, mark options={solid}]
table {%
100 274.6
300 267
500 258.6
700 254
};
\addlegendentry{180, $\flipw=2u$}
\addplot [thick, blue, mark=square*, mark size=3, mark options={solid}]
table {%
100 271.3
300 269
500 263.5
700 257
};
\addlegendentry{180, $\flipw=140$}
\addplot [thick, green01270, mark=+, mark size=3, mark options={solid}]
table {%
100 201
300 194
500 186.5
700 182.9
};
\addlegendentry{200, $\flipw=2u$}
\addplot [thick, black, mark=asterisk, mark size=3, mark options={solid}]
table {%
100 197.2
300 195.2
500 191.1
700 185.9
};
\addlegendentry{200, $\flipw=140$}
\end{axis}

\end{tikzpicture}
\caption{Examples: Comparison of $2u$ and 140 flips for instances which have 180 and 200 variables respectively. NMCTS with Dynamic SLS is better than fixed flip limit on both 180 and 200 variables type, showing that our Dynamic SLS is more adaptive to other instances with different variable numbers.}
\label{fig:orientation_fixedvsdyn_walksat_180200} 
\end{figure}

Intuitively, even if a fine tuned fixed flip limit is found for a type of instances, it is not really applicable to set as the best for other instances. However, it is obviously that along with the increasing of sizes, the flip limit should also be larger. In order to test this assumption, we proposed the dynamic SLS and showed it works well for the category 140. Therefore, in order to show the adaptation of our Dynamic SLS method, after tuning the $w$ for Dynamic SLS, we further test the the best value we get for other larger instances which have 180 and 200 variables respectively, and compare the results with the fixed flip limit way~(the best value is 140 for instances which have 140 variables). The results are presented in Fig~\ref{fig:orientation_fixedvsdyn_walksat_180200}. We see that $2u$ achieves better performance for both 180 and 200 variables categories, showing that our Dynamic SLS is more adaptive to other instances. Therefore, no redundant extra tuning cost is needed. 

\section{Experiments on Benchmark}\label{sec:full-exp}
In this section, we will show the experimental results on tested benchmark instances with aforementioned SLS based different Monte Carlo methods. The benchmark consists of 383 instances categorized by different numbers of variables. And for each category, there are a bunch of instances with different numbers of clauses.

\begin{table*}[tbh!]
\hspace{-10cm}
\scriptsize
\caption{Results of MaxSat Instances Using WalkSat based UCTMAX, NMCTS, ZNMCS and NMCS respectively, with 300 seconds budget each run, 10 repetitions each. }\label{maxsatallinstances_walksat}
\resizebox{\textwidth}{!}{\begin{tabular}{ll*7r}
\toprule
\multicolumn{2}{c}{Benchmark} & \multicolumn{6}{c}{Max\_Walksat} & \multirow{3}{*}{$\shortstack{Known\\Optimal\\Solution}$}\\
\cmidrule(r){1-2} \cmidrule(l){3-8}
Vars & Instances  &UCTMAX&NMCTS&\multicolumn{2}{c}{ZNMCS}&\multicolumn{2}{c}{NMCS}&\\
\cmidrule(lr){4-4} \cmidrule(lr){5-6} \cmidrule(lr){7-8}
 && &$m=100$&round=5,&round=1, &round=5,&round=1,&\\
 && & & level 1 & level 2 & level 1 & level 2 & \\
\midrule
70 & 50&47.7&47.8&\textbf{47.1}&47.2&\textbf{47.1}&47.7&46.8\\
80 & 50&27.3&27.4&\textbf{27.1}&\textbf{27.1}&\textbf{27.1}&27.5&26.9\\
120 & 50&223.3&219.1&219.2&\textbf{218.7}&219.0&221.0&196.1\\
140 & 50&201.8&199.2&194.0&\textbf{193.1}&195.3&195.9&184.8\\
160 & 42 &257.7&256.4&246.1&\textbf{243.1}&\textbf{243.1}&246.6&227.6\\
180&44&248.2&247.4&237.7&235.9&\textbf{235.4}&238.5&220.6\\
200& 49 &195.7&195.2&186.2&\textbf{184.5}&184.9&187.6&171.0\\
250 & 24&\textbf{7.7}&\textbf{7.7}&8.2&8.6&8.5&8.7&5.5\\
300 & 24&9.3&\textbf{9.1}&9.8&10.2&10.1&10.5&6.3\\
\bottomrule
\end{tabular}}
\end{table*}

\begin{table*}[tbh!]
\hspace{-10cm}
\scriptsize
\caption{Results of MaxSat Instances Using Novelty based UCTMAX, NMCTS, ZNMCS and NMCS respectively, with 300 seconds budget each run, 10 repetitions each. }\label{maxsatallinstances_noveltyt}
\resizebox{\textwidth}{!}{\begin{tabular}{ll*7r}
\toprule
\multicolumn{2}{c}{Benchmark} & \multicolumn{6}{c}{Max\_Novelty} & \multirow{3}{*}{$\shortstack{Known\\Optimal\\Solution}$}\\
\cmidrule(r){1-2} \cmidrule(l){3-8}
Vars & Instances  &UCTMAX&NMCTS&\multicolumn{2}{c}{ZNMCS}&\multicolumn{2}{c}{NMCS}&\\
\cmidrule(lr){4-4} \cmidrule(lr){5-6} \cmidrule(lr){7-8}
 && &m=100&round=5,&round=1, &round=5,&round=1,&\\
 && & & level 1 & level 2 & level 1 & level 2 & \\
\midrule
70 & 50&\textbf{47.1}&47.4&48.6&48.0&47.9&47.9&46.8\\
80 & 50&\textbf{27.4}&27.8&28.9&28.4&28.2&28.2&26.9\\
120 & 50&\textbf{212.7}&212.8&213.6&213.2&213.1&213.2&196.1\\
140 & 50&\textbf{185.7}&\textbf{185.7}&186.6&186.0&186.1&186.1&184.8\\
160 & 45&228.9&\textbf{228.8}&229.8&229.2&229.1&229.3&227.6\\
180 & 44&222.4&\textbf{222.2}&223.2&222.4&222.5&222.6&220.6\\
200& 49 &173.2&173.2&173.8&\textbf{173.1}&\textbf{173.1}&173.3&171.0\\
250 & 24&11.6&\textbf{11.2}&12.3&13.0&12.7&13.1&5.5\\
300 & 24&14.4&\textbf{14.0}&14.7&15.3&15.0&15.4&6.3\\
\bottomrule
\end{tabular}

}
\end{table*}

From Table~\ref{maxsatallinstances_walksat}, we can see that with WalkSAT, Nested Monte Carlo methods perform better than UCTMAX. For smaller instances like 70 and 80 variables categories, ZNMCS and NMCS level 1 perform the best, and ZNMCS level 2 achieves similar scores. Interestingly, for categories from 120 to 200, the best performance is achieved by ZNMCS level 2. And for largest instances, NMCTS is the best. These results confirm that the high level nesting of Monte Carlo methods may lead to worse performance.

From Table~\ref{maxsatallinstances_noveltyt}, we still see that for Novelty, NMCTS performs the best for larger instances. But differently, for the small instances, UCTMAX achieves best scores. Only for type 200, ZNMCS achieves the best and the scores do not vary too much. Importantly, it is clear that for most instances, Comparing with WalkSat, Novelty achieves better scores which are much more close to the known optimal solutions, which also shows that a better SLS estimate method achieves better performance together with Nested Monte Carlo. This also leads to that the improvements of NMCTS for Novelty are smaller than that for WalkSat, but we still see a possibility of increasing improvements along with the increasing of the instances sizes, which we should further investigate in future work.

In addition, the type 250 and 300 variables instances are different from others since their clauses are much more easy to be satisfied. In these cases, we find that the NMCTS performs much stably the best. 

Therefore, for both WalkSat and Novelty, we can conclude that the nesting search improves the performance of Monte Carlo methods, especially for nesting the MCTS while dealing with larger instances and employing the better SLS method.

\section{Conclusion and Future Work}

In this paper, we first investigated different rollout policies~(random, heuristics, SLS) for different Nested Monte Carlo methods, including NMCTS and NMCS to deal with MaxSAT problem. We found that heuristics are better than random, but SLS is the best rollout policy to work with Monte Carlo methods in the domain of MaxSAT. In addition, we confirmed that also for Nested Monte Carlo methods, SLS methods should also record the global best records and initialize assignment based on the found current best record. In order to further balance the exploration and exploitation, we employed $\epsilon$-greedy and found a proper $\epsilon$ value as 0.1 to randomly initialize the assignment for SLS, which improves the way that~\cite{goffinet2016monte} initialized assignment fully based on the best record. The full benchmark experimental results show that for both WalkSat and Novelty based Monte Carlo methods, the nested tree search outperforms UCTMAX~(Novelty in particularly performs better on larger instances), and NMCS with WalkSat also outperforms UCTMAX and even NMCTS. Therefore, we can conclude that nested search is important to deal with MaxSAT problems, especially for tree search on larger instances.

In the future, one way is to apply more powerful SLS algorithms together with Nested Monte Carlo methods like CCLS~\cite{luo2014ccls}. Besides, further investigation to find a light computation way for employing high level nested search is promising, especially for larger MaxSAT instances.

%\clearpage
%\newpage
%\mbox{~}
%\clearpage
\bibliographystyle{splncs04}
\bibliography{sample}

\begin{thebibliography}{10}
\providecommand{\url}[1]{\texttt{#1}}
\providecommand{\urlprefix}{URL }
\providecommand{\doi}[1]{https://doi.org/#1}

\bibitem{ansotegui2017wpm3}
Ans{\'o}tegui, C., Gabas, J.: {WPM}3: an (in) complete algorithm for weighted
  partial {MaxSAT}. Artificial Intelligence  \textbf{250},  37--57 (2017)

\bibitem{baier2012nested}
Baier, H., Winands, M.H.: Nested {M}onte {C}arlo {T}ree {S}earch for online
  planning in large mdps. In: ECAI. vol.~242, pp. 109--114 (2012)

\bibitem{brechtel2011probabilistic}
Brechtel, S., Gindele, T., Dillmann, R.: Probabilistic {MDP}-behavior planning
  for cars. In: 2011 14th International IEEE Conference on Intelligent
  Transportation Systems (ITSC). pp. 1537--1542. IEEE (2011)

\bibitem{browne2012survey}
Browne, C.B., Powley, E., Whitehouse, D., Lucas, S.M., Cowling, P.I.,
  Rohlfshagen, P., Tavener, S., Perez, D., Samothrakis, S., Colton, S.: A
  survey of {M}onte {C}arlo {T}ree {S}earch methods. IEEE Transactions on
  Computational Intelligence and AI in games  \textbf{4}(1),  1--43 (2012)

\bibitem{cazenave2009nested}
Cazenave, T.: Nested monte-carlo search. In: Twenty-First International Joint
  Conference on Artificial Intelligence (2009)

\bibitem{cazenave2020generalized}
Cazenave, T.: Generalized nested rollout policy adaptation. In: {M}onte {C}arlo
  Search International Workshop. pp. 71--83. Springer (2020)

\bibitem{cazenave2012application}
Cazenave, T., Teytaud, F.: Application of the nested rollout policy adaptation
  algorithm to the traveling salesman problem with time windows. In:
  International Conference on Learning and Intelligent Optimization. pp.
  42--54. Springer (2012)

\bibitem{cowling2012ensemble}
Cowling, P.I., Ward, C.D., Powley, E.J.: Ensemble determinization in {M}onte
  {C}arlo {T}ree {S}earch for the imperfect information card game magic: The
  gathering. IEEE Transactions on Computational Intelligence and AI in Games
  \textbf{4}(4),  241--257 (2012)

\bibitem{demaine2006morpion}
Demaine, E.D., Demaine, M.L., Langerman, A., Langerman, S.: Morpion
  {S}olitaire. Theory of Computing Systems  \textbf{39}(3),  439--453 (2006)

\bibitem{douxdeep}
Doux, B., Negrevergne, B., Cazenave, T.: Deep reinforcement learning for
  {M}orpion {S}olitaire. In: Advances in Computer Games (2021)

\bibitem{gelly2007combining}
Gelly, S., Silver, D.: Combining online and offline knowledge in {UCT}. In:
  Proceedings of the 24th international conference on Machine learning. pp.
  273--280 (2007)

\bibitem{goffinet2016monte}
Goffinet, J., Ramanujan, R.: Monte-carlo tree search for the maximum
  satisfiability problem. In: International Conference on Principles and
  Practice of Constraint Programming. pp. 251--267. Springer (2016)

\bibitem{heras2008minimaxsat}
Heras, F., Larrosa, J., Oliveras, A.: Mini{MaxSAT}: {A}n efficient weighted
  {MaxSAT} solver. Journal of Artificial Intelligence Research  \textbf{31},
  1--32 (2008)

\bibitem{ignatiev2019rc2}
Ignatiev, A., Morgado, A., Marques-Silva, J.: {RC}2: an efficient {M}ax{SAT}
  solver. Journal on Satisfiability, Boolean Modeling and Computation
  \textbf{11}(1),  53--64 (2019)

\bibitem{kautz2004walksat}
Kautz, H., Selman, B., McAllester, D.: Walksat in the 2004 {SAT} {C}ompetition.
  In: Proceedings of the International Conference on Theory and Applications of
  Satisfiability Testing (2004)

\bibitem{kroc2009integrating}
Kroc, L., Sabharwal, A., Gomes, C.P., Selman, B.: Integrating systematic and
  local search paradigms: A new strategy for {MaxSAT}. In: Twenty-First
  International Joint Conference on Artificial Intelligence (2009)

\bibitem{luo2014ccls}
Luo, C., Cai, S., Wu, W., Jie, Z., Su, K.: {CCLS}: an efficient local search
  algorithm for weighted maximum satisfiability. IEEE Transactions on Computers
   \textbf{64}(7),  1830--1843 (2014)

\bibitem{martins2014open}
Martins, R., Manquinho, V., Lynce, I.: Open-{WBO}: A modular {M}ax{SAT} solver.
  In: International Conference on Theory and Applications of Satisfiability
  Testing. pp. 438--445. Springer (2014)

\bibitem{menai2003efficient}
Menai, M.E.b., Batouche, M.: Efficient initial solution to extremal
  optimization algorithm for weighted {MAXSAT} problem. In: International
  Conference on Industrial, Engineering and Other Applications of Applied
  Intelligent Systems. pp. 592--603. Springer (2003)

\bibitem{morgado2013iterative}
Morgado, A., Heras, F., Liffiton, M., Planes, J., Marques-Silva, J.: Iterative
  and core-guided {MaxSAT} solving: {A} survey and assessment. Constraints
  \textbf{18}(4),  478--534 (2013)

\bibitem{pelikan2003hierarchical}
Pelikan, M., Goldberg, D.E.: Hierarchical {BOA} solves ising spin glasses and
  {MAXSAT}. In: Genetic and Evolutionary Computation Conference. pp.
  1271--1282. Springer (2003)

\bibitem{rosin2011nested}
Rosin, C.D.: Nested rollout policy adaptation for {M}onte {C}arlo {T}ree
  {S}earch. In: Twenty-Second International Joint Conference on Artificial
  Intelligence (2011)

\bibitem{seify2020single}
Seify, A., Buro, M.: Single-agent optimization through policy iteration using
  {M}onte {C}arlo {T}ree {S}earch. arXiv preprint arXiv:2005.11335  (2020)

\bibitem{tompkins2004ubcsat}
Tompkins, D.A., Hoos, H.H.: {UBCSAT}: {A}n implementation and experimentation
  environment for {SLS} algorithms for {SAT} and {MAXSAT}. In: International
  conference on theory and applications of satisfiability testing. pp.
  306--320. Springer (2004)

\bibitem{wang2018assessing}
Wang, H., Emmerich, M., Plaat, A.: Assessing the potential of classical
  {Q}-learning in {G}eneral {G}ame {P}laying. In: Benelux Conference on
  Artificial Intelligence. pp. 138--150. Springer (2018)

\bibitem{wang2020analysis}
Wang, H., Emmerich, M., Preuss, M., Plaat, A.: Analysis of hyper-parameters for
  small games: Iterations or epochs in self-play? arXiv preprint
  arXiv:2003.05988  (2020)

\bibitem{wang2020tackling}
Wang, H., Preuss, M., Emmerich, M., Plaat, A.: Tackling {M}orpion {S}olitaire
  with {A}lpha{Z}ero-like ranked reward reinforcement learning. In: 2020 22nd
  International Symposium on Symbolic and Numeric Algorithms for Scientific
  Computing (SYNASC). pp. 149--152. IEEE (2020)

\bibitem{wang2020warm}
Wang, H., Preuss, M., Plaat, A.: Warm-{S}tart {A}lpha{Z}ero self-play search
  enhancements. In: Proceedings of the Parallel Problem Solving from Nature –
  PPSN XVI. pp. 528--542 (2020)

\bibitem{wang2021adaptive}
Wang, H., Preuss, M., Plaat, A.: Adaptive warm-start {MCTS} in
  {A}lpha{Z}ero-like deep reinforcement learning. In: Pacific Rim International
  Conference on Artificial Intelligence. pp. 60--71. Springer (2021)

\end{thebibliography}

\newpage
\section*{Appendix}

\subsection{Detail Results for Each Instances and Each Categories}
\begin{table*}[tbh!]
\hspace{-10cm}
\scriptsize
\caption{Results of Max3Sat~(70 variables) on different Instances Using SingleUCT SLS, MultiUCT SLS and NMC SLS respectively, 10 repetitions each. }\label{max3sat70variable_sls}
\resizebox{\textwidth}{!}{\begin{tabular}{ll*{7}{r}}
\toprule
Clauses & Ins &$UCTMax$&$NMCTSMax$&\multicolumn{2}{c}{$NMCSZMax$}&\multicolumn{2}{c}{$NMCSMax$}&Optimal\\
&&&&&&&& Solution\\
\cmidrule(lr){5-6}\cmidrule(lr){7-8}
 & & &m=100&rounds=5&rounds=1&rounds=5&rounds=1& \\
       &     & &     &level 1&level 2&level 1&level 2& \\
\midrule
\multirow{10}{*}{800}
&1&769.0&769.0&769.0&769.0&769.0&769.0&769\\
&2&766.0&766.0&765.9&766.0&766.0&765.8&766\\
&3&770.0&769.9&770.0&770.0&770.0&770.0&770\\
&4&772.0&772.0&772.0&772.0&772.0&772.0&772\\
&5&768.1&768.8&768.9&769.0&768.9&768.7&769\\
&6&770.0&770.0&770.0&770.0&770.0&770.0&770\\
&7&768.9&768.9&769.0&768.9&768.9&768.9&769\\
&8&765.9&766.0&766.0&766.0&766.0&766.0&766\\
&9&767.7&767.8&768.0&768.0&767.9&767.3&768\\
&10&769.7&770.0&769.9&769.9&770.0&769.4&770\\
\midrule
\multirow{10}{*}{900}
&1&861.0&861.0&861.0&861.0&861.0&860.7&861\\
&2&861.4&861.6&862.0&862.0&862.0&861.8&862\\
&3&860.8&860.3&861.0&861.0&861.0&860.9&861\\
&4&860.4&860.4&861.0&861.0&861.0&860.8&861\\
&5&859.8&859.8&859.9&860.0&860.0&859.9&860\\
&6&859.0&859.0&859.0&859.0&859.0&858.9&859\\
&7&859.6&859.0&860.0&860.0&860.0&859.9&860\\
&8&858.0&858.0&858.0&858.0&858.0&858.0&858\\
&9&865.0&865.0&865.0&865.0&865.0&865.0&865\\
&10&861.0&861.0&861.0&861.0&861.0&860.8&861\\
\midrule
\multirow{10}{*}{1000}
&1&953.0&953.0&953.0&953.0&953.0&953.0&953\\
&2&956.8&957.0&957.0&957.0&957.0&957.0&957\\
&3&955.0&955.0&955.0&955.0&955.0&955.0&955\\
&4&952.3&951.5&953.0&953.0&953.0&952.6&953\\
&5&958.0&958.0&958.0&958.0&958.0&958.0&958\\
&6&949.8&949.5&949.1&949.1&949.3&949.0&950\\
&7&950.8&951.0&951.0&951.0&951.0&950.8&951\\
&8&951.7&951.9&952.0&952.0&952.0&951.9&952\\
&9&949.8&949.6&950.7&950.5&950.5&950.4&951\\
&10&955.0&954.9&955.0&955.0&955.0&955.0&955\\
\midrule
\multirow{10}{*}{1100}
&1&1043.9&1044.0&1044.0&1044.0&1044.0&1044.0&1044\\
&2&1044.0&1043.4&1044.9&1044.9&1044.9&1044.8&1045\\
&3&1045.6&1046.0&1046.3&1046.8&1046.3&1046.2&1047\\
&4&1047.8&1047.7&1047.9&1048.0&1048.0&1047.9&1048\\
&5&1046.2&1046.2&1046.7&1047.0&1046.9&1046.6&1047\\
&6&1046.9&1046.1&1046.4&1047.0&1047.0&1046.6&1047\\
&7&1046.2&1046.3&1046.9&1046.9&1047.0&1046.8&1047\\
&8&1048.9&1048.9&1049.0&1049.0&1049.0&1048.9&1049\\
&9&1052.0&1052.0&1052.0&1052.0&1052.0&1052.0&1052\\
&10&1041.9&1041.9&1041.9&1042.0&1042.0&1041.9&1042\\
\midrule
\multirow{10}{*}{1200}
&1&1133.1&1132.9&1133.7&1133.9&1134.0&1133.8&1134\\
&2&1135.5&1135.5&1136.9&1136.6&1136.3&1135.5&1137\\
&3&1134.0&1134.6&1134.7&1135.0&1135.0&1134.8&1135\\
&4&1131.9&1131.7&1132.6&1132.9&1132.8&1132.8&1133\\
&5&1134.5&1134.4&1134.3&1134.6&1134.5&1134.3&1135\\
&6&1133.0&1133.0&1133.8&1134.0&1133.9&1133.8&1134\\
&7&1137.2&1137.9&1138.0&1137.9&1138.0&1137.9&1138\\
&8&1135.1&1133.8&1136.6&1136.7&1137.0&1136.9&1137\\
&9&1138.8&1139.0&1139.0&1139.0&1139.0&1138.8&1139\\
&10&1136.1&1136.7&1137.0&1136.8&1137.0&1136.7&1137\\

\midrule
\multicolumn{2}{l}{Aggregate}&952.8&952.7&953.1&953.1&953.1&953.0&953.18\\
\bottomrule
\end{tabular}
}
\end{table*}

%\begin{table}[tbh!]
%\hspace{-10cm}
%\scriptsize
%\caption{Results of Mastering Max3Sat~(80 variables) on different Instances Using SingleUCT SLS, MultiUCT SLS and NMC SLS respectively, 10 repetitions each. }\label{max3sat80variable_sls}
%\resizebox{\textwidth}{!}{\input{tables/fullexperiments_80v}}
%\end{table}

\begin{table*}[tbh!]
\hspace{-10cm}
\scriptsize
\caption{Results of Max2Sat~(120 variables) on different Instances Using SingleUCT SLS, MultiUCT SLS and NMC SLS respectively, 10 repetitions each. }\label{max2sat120variable_sls}
\resizebox{\textwidth}{!}{\begin{tabular}{ll*{7}{r}}
\toprule
Clauses & Ins &$UCTMax$&$NMCTSMax$&\multicolumn{2}{c}{$NMCSZMax$}&\multicolumn{2}{c}{$NMCSMax$}&Optimal\\
&&&&&&&& Solution\\
\cmidrule(lr){5-6}\cmidrule(lr){7-8}
 & & &m=100&rounds=5&rounds=1&rounds=5&rounds=1& \\
       &     & &     &level 1&level 2&level 1&level 2& \\
\midrule
\multirow{10}{*}{1200}
&1&1026.0&1025.8&1032.3&1032.0&1032.0&1032.1&1039\\
&2&1026.6&1027.5&1033.9&1034.2&1033.5&1032.7&1041\\
&3&1024.1&1024.4&1031.0&1033.8&1034.0&1033.0&1040\\
&4&1029.8&1029.9&1031.0&1037.2&1036.6&1036.6&1043\\
&5&1046.9&1045.8&1051.8&1051.4&1050.5&1051.9&1057\\
&6&1019.4&1020.3&1024.5&1025.0&1024.1&1024.8&1033\\
&7&1022.8&1023.7&1029.9&1030.6&1030.7&1027.6&1038\\
&8&1019.6&1020.6&1027.8&1029.0&1027.8&1028.6&1035\\
&9&1041.1&1042.3&1047.7&1047.4&1049.2&1047.0&1052\\
&10&1034.0&1034.2&1040.4&1041.3&1041.3&1040.8&1046\\
\midrule
\multirow{10}{*}{1300}
&1&1105.7&1106.3&1111.0&1113.4&1113.4&1112.1&1120\\
&2&1115.6&1115.0&1119.6&1121.7&1121.0&1120.7&1128\\
&3&1112.2&1112.4&1118.6&1120.4&1117.4&1119.6&1127\\
&4&1109.6&1109.6&1113.8&1116.0&1118.1&1115.9&1124\\
&5&1118.6&1118.9&1124.0&1125.1&1125.4&1123.8&1132\\
&6&1106.9&1106.6&1111.3&1111.8&1113.9&1113.1&1120\\
&7&1114.9&1114.6&1124.0&1124.6&1124.3&1121.5&1131\\
&8&1109.3&1109.7&1115.5&1118.1&1117.0&1116.3&1126\\
&9&1099.8&1099.8&1105.3&1108.2&1107.5&1106.3&1114\\
&10&1108.9&1107.6&1112.2&1113.6&1114.6&1112.0&1120\\
\midrule
\multirow{10}{*}{1400}
&1&1187.2&1187.2&1192.4&1196.1&1196.0&1192.7&1203\\
&2&1195.2&1193.7&1198.1&1200.8&1201.3&1199.0&1209\\
&3&1195.7&1195.9&1197.4&1202.4&1201.6&1200.6&1211\\
&4&1181.7&1181.8&1185.6&1189.5&1189.7&1186.6&1200\\
&5&1185.0&1184.3&1191.0&1194.3&1192.8&1189.8&1201\\
&6&1186.8&1188.8&1191.6&1196.4&1196.6&1195.8&1204\\
&7&1176.2&1175.8&1180.4&1185.1&1184.0&1182.1&1194\\
&8&1192.3&1192.6&1196.0&1198.2&1197.2&1196.4&1206\\
&9&1185.9&1186.5&1189.9&1194.5&1193.8&1191.2&1202\\
&10&1171.5&1171.5&1172.6&1177.2&1177.5&1176.6&1189\\
\midrule
\multirow{10}{*}{1500}
&1&1274.3&1274.2&1279.6&1282.6&1282.5&1281.8&1289\\
&2&1269.8&1270.8&1273.4&1278.8&1277.7&1276.4&1287\\
&3&1275.9&1277.1&1279.4&1283.6&1284.3&1280.6&1293\\
&4&1271.5&1271.7&1274.6&1277.1&1278.9&1278.8&1288\\
&5&1248.5&1248.8&1251.1&1256.9&1257.6&1253.9&1267\\
&6&1273.6&1272.2&1277.4&1281.6&1282.5&1279.1&1291\\
&7&1269.0&1269.0&1272.2&1277.3&1275.1&1274.4&1284\\
&8&1271.7&1270.4&1275.3&1279.4&1280.1&1277.9&1288\\
&9&1262.0&1261.7&1263.9&1269.4&1267.4&1268.9&1277\\
&10&1269.0&1269.2&1270.7&1275.3&1276.3&1276.8&1287\\
\midrule
\multirow{10}{*}{1600}
&1&1274.3&1274.2&1279.6&1282.6&1282.5&1281.8&1367\\
&2&1269.8&1270.8&1273.4&1278.8&1277.7&1276.4&1361\\
&3&1275.9&1277.1&1279.4&1283.6&1284.3&1280.6&1367\\
&4&1271.5&1271.7&1274.6&1277.1&1278.9&1278.8&1381\\
&5&1248.5&1248.8&1251.1&1256.9&1257.6&1253.9&1353\\
&6&1273.6&1272.2&1277.4&1281.6&1282.5&1279.1&1365\\
&7&1269.0&1269.0&1272.2&1277.3&1275.1&1274.4&1375\\
&8&1271.7&1270.4&1275.3&1279.4&1280.1&1277.9&1363\\
&9&1262.0&1261.7&1263.9&1269.4&1267.4&1268.9&1360\\
&10&1269.0&1269.2&1270.7&1275.3&1276.3&1276.8&1367\\

\midrule
\multicolumn{2}{l}{Aggregate}&1172.4&1172.5&1176.7&1179.9&1179.8&1178.5&1203.9\\
\bottomrule
\end{tabular}}
\end{table*}

\begin{table*}[tbh!]
\hspace{-10cm}
\scriptsize
\caption{Results of Max2Sat~(140 variables) on different Instances Using SingleUCT SLS, MultiUCT SLS and NMC SLS respectively, 10 repetitions each. }\label{max2sat140variable_sls}
\resizebox{\textwidth}{!}{\begin{tabular}{ll*{7}{r}}
\toprule
Clauses & Ins &$UCTMax$&$NMCTSMax$&\multicolumn{2}{c}{$NMCSZMax$}&\multicolumn{2}{c}{$NMCSMax$}&Optimal\\
&&&&&&&& Solution\\
\cmidrule(lr){5-6}\cmidrule(lr){7-8}
 & & &m=100&rounds=5&rounds=1&rounds=5&rounds=1& \\
       &     & &     &level 1&level 2&level 1&level 2& \\
\midrule
\multirow{10}{*}{1200}
&1&1040.2&1041.0&1043.7&1046.0&1048.2&1046.6&1056\\
&2&1030.6&1030.4&1032.1&1035.8&1035.8&1036.3&1045\\
&3&1029.6&1029.3&1032.7&1035.6&1034.3&1036.0&1045\\
&4&1038.0&1038.5&1040.6&1042.2&1044.1&1040.7&1052\\
&5&1044.1&1044.5&1047.2&1050.0&1050.0&1049.3&1057\\
&6&1037.9&1037.6&1040.3&1044.6&1044.0&1042.6&1052\\
&7&1036.2&1036.1&1037.8&1043.7&1043.6&1040.8&1052\\
&8&1032.1&1033.7&1036.3&1040.0&1039.5&1037.0&1048\\
&9&1034.9&1035.4&1037.1&1041.4&1041.0&1041.5&1049\\
&10&1046.1&1046.2&1047.5&1050.9&1052.0&1051.5&1060\\
&1&1120.9&1121.0&1124.8&1128.2&1128.6&1128.2&1138\\
&2&1112.8&1113.8&1116.5&1120.6&1119.9&1118.4&1129\\
&3&1115.9&1116.4&1118.5&1123.6&1123.6&1122.4&1132\\
&4&1118.1&1118.5&1123.3&1126.8&1125.3&1126.3&1136\\
&5&1113.5&1114.9&1115.8&1122.3&1120.7&1120.6&1131\\
&6&1115.8&1116.2&1118.5&1123.2&1124.3&1123.0&1132\\
&7&1125.9&1125.7&1127.7&1132.5&1132.1&1132.0&1140\\
&8&1128.8&1129.1&1130.2&1136.7&1136.4&1134.6&1143\\
&9&1123.1&1122.4&1124.8&1129.7&1129.0&1127.8&1138\\
&10&1111.6&1111.9&1113.4&1119.9&1118.6&1117.4&1130\\
&1&1200.4&1200.2&1203.0&1207.5&1207.2&1208.7&1218\\
&2&1202.5&1201.7&1204.6&1210.8&1213.1&1208.0&1222\\
&3&1187.9&1187.3&1191.2&1198.0&1196.6&1194.8&1207\\
&4&1197.4&1198.0&1200.5&1207.6&1207.6&1206.5&1216\\
&5&1195.2&1194.5&1196.9&1202.1&1203.5&1202.7&1213\\
&6&1195.5&1195.9&1198.4&1200.8&1202.3&1202.1&1212\\
&7&1192.4&1192.3&1195.1&1200.4&1202.0&1198.9&1213\\
&8&1198.5&1200.8&1202.7&1206.4&1207.4&1205.8&1219\\
&9&1197.4&1198.9&1202.3&1206.3&1206.1&1202.8&1215\\
&10&1190.5&1191.6&1192.6&1199.3&1200.0&1199.7&1212\\
&1&1276.9&1277.1&1280.3&1285.1&1283.4&1284.5&1295\\
&2&1284.4&1282.9&1283.5&1289.8&1293.1&1290.2&1301\\
&3&1266.9&1267.8&1270.1&1275.2&1276.5&1273.8&1288\\
&4&1281.1&1281.3&1283.7&1291.2&1292.0&1289.6&1303\\
&5&1274.0&1274.6&1277.4&1283.2&1282.6&1281.9&1295\\
&6&1281.4&1283.2&1285.5&1290.0&1290.0&1289.5&1302\\
&7&1278.9&1279.1&1281.4&1285.8&1286.2&1285.1&1298\\
&8&1279.7&1281.8&1284.2&1289.8&1290.3&1289.9&1301\\
&9&1282.0&1282.1&1284.7&1288.6&1288.0&1288.0&1301\\
&10&1278.4&1278.6&1281.4&1287.8&1286.9&1286.3&1298\\
&1&1357.9&1359.4&1362.6&1368.7&1368.0&1366.4&1379\\
&2&1358.5&1359.4&1362.0&1367.3&1364.7&1362.3&1379\\
&3&1350.7&1351.3&1352.8&1358.9&1360.0&1357.9&1374\\
&4&1357.7&1358.3&1361.8&1366.9&1361.8&1365.4&1380\\
&5&1350.1&1351.5&1353.2&1359.2&1358.7&1357.0&1372\\
&6&1357.9&1356.3&1360.4&1362.9&1367.8&1366.1&1380\\
&7&1361.4&1362.3&1364.1&1368.8&1369.9&1366.8&1382\\
&8&1351.1&1352.0&1352.3&1359.5&1360.7&1359.6&1373\\
&9&1351.4&1350.1&1352.1&1358.9&1356.6&1357.2&1372\\
&10&1353.0&1354.6&1357.2&1363.0&1361.2&1361.8&1374\\
statistic&1196.9&1197.4&1199.7&1204.7&1204.7&1203.6&1215.18\\

\midrule
\multicolumn{2}{l}{Aggregate}&1196.9&1197.4&1199.7&1204.7&1204.7&1203.6&1215.18\\
\bottomrule
\end{tabular}
}
\end{table*}

\begin{table*}[tbh!]
\hspace{-10cm}
\scriptsize
\caption{Results of Max2Sat~(200 variables) on different Instances Using SingleUCT SLS, MultiUCT SLS and NMC SLS respectively, 10 repetitions each. }\label{max2sat200variable_walksat}
\resizebox{\textwidth}{!}{\begin{tabular}{ll*{7}{r}}
\toprule
Clauses & Ins &$UCTMax$&$NMCTSMax$&\multicolumn{2}{c}{$NMCSZMax$}&\multicolumn{2}{c}{$NMCSMax$}&Optimal\\
&&&&&&&& Solution\\
\cmidrule(lr){5-6}\cmidrule(lr){7-8}
 & & &m=100&rounds=5&rounds=1&rounds=5&rounds=1& \\
       &     & &     &level 1&level 2&level 1&level 2& \\
\midrule
\multirow{7}{*}{1200}
&1&1065.9&1065.9&1067.5&1070.3&1069.8&1070.3&1082\\
&2&1055.3&1055.3&1058.3&1062.0&1060.4&1060.9&1073\\
&3&1058.3&1059.1&1059.9&1063.2&1063.7&1063.4&1075\\
&4&1069.0&1068.9&1071.4&1075.6&1077.9&1077.5&1085\\
&5&1066.1&1066.6&1067.7&1072.0&1073.7&1072.5&1083\\
&6&1054.0&1053.9&1056.4&1060.4&1059.0&1058.7&1073\\
&7&1072.5&1073.3&1075.2&1080.3&1079.3&1079.8&1088\\
&1&1136.6&1135.8&1136.2&1141.0&1139.8&1140.4&1156\\
&2&1135.1&1136.9&1137.7&1143.0&1142.5&1141.7&1157\\
&3&1161.6&1162.1&1163.8&1167.4&1167.5&1166.6&1179\\
&4&1141.2&1141.3&1142.9&1146.6&1147.8&1145.0&1161\\
&5&1136.9&1137.9&1138.9&1142.8&1142.6&1142.5&1158\\
&6&1147.4&1146.4&1148.3&1152.7&1153.1&1152.7&1166\\
&7&1156.3&1156.8&1158.5&1162.1&1161.5&1160.6&1172\\
&1&1222.5&1221.6&1224.0&1231.4&1230.3&1229.7&1244\\
&2&1233.7&1235.2&1235.8&1240.9&1239.6&1237.1&1253\\
&3&1227.8&1227.4&1229.3&1233.3&1234.3&1235.0&1250\\
&4&1213.7&1213.4&1214.8&1221.1&1220.5&1218.9&1236\\
&5&1225.7&1225.6&1227.1&1232.1&1230.9&1231.4&1247\\
&6&1237.0&1236.9&1239.4&1244.0&1241.9&1244.6&1255\\
&7&1240.4&1240.1&1241.6&1245.2&1244.9&1243.9&1258\\
&1&1313.5&1313.8&1313.8&1319.0&1317.7&1320.1&1335\\
&2&1313.3&1311.9&1315.1&1319.1&1317.5&1319.6&1335\\
&3&1311.8&1312.3&1315.1&1320.0&1318.3&1320.0&1335\\
&4&1297.6&1298.9&1300.5&1305.2&1305.5&1302.7&1322\\
&5&1302.5&1303.0&1303.8&1307.8&1308.0&1307.0&1326\\
&6&1297.7&1295.7&1298.7&1302.7&1302.7&1304.4&1320\\
&7&1294.5&1295.7&1296.0&1299.6&1302.5&1301.6&1321\\
&1&1375.7&1376.9&1379.0&1381.9&1382.9&1381.4&1405\\
&2&1398.8&1399.1&1401.8&1406.6&1405.7&1404.0&1422\\
&3&1387.1&1387.8&1389.7&1394.4&1395.3&1394.5&1414\\
&4&1376.9&1377.4&1380.3&1384.1&1386.7&1384.5&1405\\
&5&1377.6&1379.2&1380.1&1386.2&1384.0&1385.2&1406\\
&6&1398.0&1399.9&1401.1&1406.1&1405.8&1403.3&1421\\
&7&1380.7&1382.4&1384.5&1387.4&1388.6&1381.4&1408\\
&1&1479.3&1480.1&1483.0&1488.3&1488.0&1490.6&1507\\
&2&1465.8&1465.5&1466.5&1475.3&1470.9&1473.7&1493\\
&3&1458.5&1459.4&1460.3&1466.2&1464.1&1467.1&1487\\
&4&1460.2&1461.0&1463.3&1469.0&1468.1&1468.0&1488\\
&5&1456.2&1455.3&1459.3&1463.1&1462.9&1461.5&1483\\
&6&1462.3&1463.9&1465.0&1471.7&1472.0&1470.0&1491\\
&7&1470.7&1471.2&1474.9&1479.4&1477.7&1479.9&1499\\
&1&1565.0&1564.5&1568.3&1572.1&1566.1&1564.9&1594\\
&2&1546.8&1547.9&1549.5&1552.7&1555.4&1555.8&1577\\
&3&1542.1&1541.3&1542.5&1549.1&1548.0&1546.6&1571\\
&4&1547.4&1547.9&1550.7&1554.0&1555.8&1553.7&1579\\
&5&1545.8&1546.4&1547.6&1557.0&1558.1&1555.4&1580\\
&6&1539.5&1542.9&1544.5&1550.2&1547.8&1547.8&1572\\
&7&1544.6&1543.1&1544.9&1551.7&1550.6&1551.7&1574\\

\midrule
\multicolumn{2}{l}{Aggregate}&1305.4&1305.8&1307.6&1312.4&1312.0&1311.6&1329.0\\
\bottomrule
\end{tabular}}
\end{table*}

\begin{table*}[tbh!]
\hspace{-10cm}
\scriptsize
\caption{Results of Max2Sat~(250 variables) on different Instances Using SingleUCT SLS, MultiUCT SLS and NMC SLS respectively, 10 repetitions each. }\label{max2sat250variable_walksat}
\resizebox{\textwidth}{!}{\begin{tabular}{ll*{7}{r}}
\toprule
Clauses & Ins &$UCTMax$&$NMCTSMax$&\multicolumn{2}{c}{$NMCSZMax$}&\multicolumn{2}{c}{$NMCSMax$}&Optimal\\
&&&&&&&& Solution\\
\cmidrule(lr){5-6}\cmidrule(lr){7-8}
 & & &m=100&rounds=5&rounds=1&rounds=5&rounds=1& \\
       &     & &     &level 1&level 2&level 1&level 2& \\
\midrule
\multirow{24}{*}{1000}
&1&992.7&992.1&992.2&991.6&991.5&991.0&995\\
&2&992.9&992.8&992.0&991.7&992.6&992.3&995\\
&3&992.2&992.2&992.4&991.3&991.6&991.4&995\\
&4&991.8&992.6&992.7&991.4&991.0&991.5&994\\
&5&992.4&991.8&991.1&991.1&990.8&990.5&994\\
&6&992.3&992.6&991.6&991.0&991.2&991.6&994\\
&7&992.4&992.1&991.2&990.7&991.5&990.9&994\\
&8&992.7&992.5&992.0&991.4&991.9&991.4&995\\
&9&991.5&991.9&991.3&990.8&991.0&991.0&994\\
&10&992.2&992.4&992.2&991.7&992.4&991.5&994\\
&11&992.4&992.5&991.8&992.0&991.6&991.2&994\\
&12&991.5&991.7&991.7&991.0&991.1&991.3&994\\
&13&993.0&992.6&991.5&991.3&990.8&991.9&995\\
&14&992.4&992.7&991.7&991.7&991.9&991.3&995\\
&15&992.7&993.5&991.8&993.0&992.1&992.0&995\\
&16&992.2&991.8&991.4&991.0&990.9&991.0&995\\
&17&991.7&991.9&990.9&990.5&991.1&990.6&994\\
&18&991.8&991.7&991.4&990.4&990.5&991.0&994\\
&19&993.5&993.4&992.7&991.5&991.9&991.8&995\\
&20&991.5&991.1&991.4&990.9&991.1&990.5&993\\
&21&992.4&992.1&991.6&990.7&991.2&991.1&996\\
&22&992.0&992.3&991.9&991.4&991.7&991.3&995\\
&23&992.7&992.6&992.0&992.1&991.6&991.8&994\\
&24&993.4&992.9&992.2&993.1&992.8&991.9&995\\
\midrule
\multicolumn{2}{l}{Aggregate}&992.3&992.3&991.8&991.4&991.5&991.3&994.5\\
\bottomrule
\end{tabular}}
\end{table*}

\begin{table*}[tbh!]
\hspace{-10cm}
\scriptsize
\caption{Results of Max2Sat~(300 variables) on different Instances Using SingleUCT SLS, MultiUCT SLS and NMC SLS respectively, 10 repetitions each. }\label{max2sat300variable_walksat}
\resizebox{\textwidth}{!}{\begin{tabular}{ll*{7}{r}}
\toprule
Clauses & Ins &$UCTMax$&$NMCTSMax$&\multicolumn{2}{c}{$NMCSZMax$}&\multicolumn{2}{c}{$NMCSMax$}&Optimal\\
&&&&&&&& Solution\\
\cmidrule(lr){5-6}\cmidrule(lr){7-8}
 & & &m=100&rounds=5&rounds=1&rounds=5&rounds=1& \\
       &     & &     &level 1&level 2&level 1&level 2& \\
\midrule
\multirow{24}{*}{1200}
&1&1191.2&1190.7&1190.7&1190.7&1189.7&1190.8&1194\\
&2&1191.0&1190.8&1190.4&1189.9&1190.6&1189.8&1195\\
&3&1191.0&1190.9&1190.4&1189.5&1189.2&1190.0&1194\\
&4&1190.6&1190.4&1189.6&1189.7&1189.5&1188.8&1193\\
&5&1190.7&1191.1&1190.5&1189.9&1190.4&1189.5&1194\\
&6&1190.8&1191.3&1191.0&1190.0&1190.4&1189.6&1194\\
&7&1191.6&1191.9&1190.9&1191.1&1190.9&1190.1&1195\\
&8&1189.9&1190.4&1189.8&1189.6&1189.5&1188.8&1193\\
&9&1190.4&1190.7&1189.9&1189.2&1189.7&1188.4&1193\\
&10&1190.8&1190.7&1189.7&1190.1&1189.8&1189.6&1193\\
&11&1191.1&1191.9&1190.5&1189.7&1190.6&1190.1&1194\\
&12&1190.7&1190.8&1190.0&1189.9&1189.9&1189.4&1193\\
&13&1191.4&1191.6&1190.8&1190.2&1190.3&1190.8&1194\\
&14&1191.0&1190.6&1190.9&1189.7&1189.8&1189.3&1194\\
&15&1190.4&1191.1&1189.8&1189.2&1189.6&1188.7&1193\\
&16&1190.0&1190.8&1190.1&1189.5&1189.9&1189.8&1194\\
&17&1190.6&1190.4&1190.4&1189.5&1188.8&1189.7&1193\\
&18&1191.3&1190.9&1190.0&1190.2&1189.7&1189.9&1193\\
&19&1190.4&1190.3&1189.5&1189.5&1189.8&1188.9&1194\\
&20&1190.1&1189.9&1189.3&1189.5&1189.1&1189.3&1194\\
&21&1190.0&1190.7&1189.4&1189.3&1189.2&1188.8&1195\\
&22&1191.2&1191.2&1190.4&1189.7&1189.9&1189.9&1194\\
&23&1190.3&1191.3&1190.5&1190.0&1190.9&1189.2&1193\\
&24&1191.4&1190.9&1190.3&1189.7&1190.3&1189.8&1193\\

\midrule
\multicolumn{2}{l}{Aggregate}&1190.7&1190.9&1190.2&1189.8&1189.9&1189.5&1193.7\\
\bottomrule
\end{tabular}}
\end{table*}

\end{document}